\documentclass[runningheads]{llncs}
\usepackage[T1]{fontenc}

\usepackage{multirow}      
\usepackage{pifont}
\usepackage{graphicx}
\usepackage{amsmath}
\usepackage{amssymb}
\usepackage{booktabs}
\usepackage{multirow}
\usepackage{makecell}

\usepackage{xcolor}
\usepackage{colortbl}
\definecolor{lightergray}{cmyk}{0,0.0,0.0,0.14}
\usepackage{graphicx}
\begin{document}

\title{H3D-MarNet: Wavelet-Guided Dual-Path Learning for Metal Artifact Suppression and CT Modality Transformation for Radiotherapy Workflows}
\titlerunning{H3D-MarNet}

\author{Mubashara Rehman\inst{1,2}\orcidID{0009-0007-2935-0409} \and
Niki Martinel\inst{1}\orcidID{0000-0002-6962-8643} \and
Michele Avanzo\inst{2}\orcidID{0000-0003-1711-4242} \and
Riccardo Spizzo\inst{2}\orcidID{0000-0001-7772-0960} \and
Christian Micheloni\inst{1}\orcidID{0000-0003-4503-7483}}

\authorrunning{R.Mubashara et al.}                                        

\institute{
Machine Learning and Perception Lab, Università degli Studi di Udine, 33100 Udine, Italy\\
\email{rehman.mubashara@spes.uniud.it}\\
\email{\{niki.martinel, christian.micheloni\}@uniud.it}
\and
Centro di Riferimento Oncologico di Aviano IRCCS, 33081 Aviano, Italy\\
\email{\{mavanzo, rspizzo\}@cro.it}
}



\maketitle              

\begin{abstract}

Metal artifacts in computed tomography (CT) severely degrade image quality, compromising diagnostic accuracy and radiotherapy planning, especially in cancer patients with high-density implants. We propose H3D-MarNet, a two-stage framework for artifact-aware CT domain transformation from kilo-voltage CT (kVCT) to mega-voltage CT (MVCT). In the first stage, a wavelet-based preprocessing module suppresses metal-induced artifacts through frequency-aware denoising while preserving anatomical structures. In second stage, Domain-TransNet performs kVCT-to-MVCT domain transformation using a hybrid volumetric learning architecture. Domain-TransNet integrates a CNN-based encoder to capture fine-grained local anatomical details and a transformer-based encoder to model long-range volumetric dependencies. The complementary representations are fused through an attention-based feature fusion mechanism to ensure spatial and contextual coherence across slices. A multi-stage, attention-guided decoder, supported by deep supervision, progressively reconstructs artifact-suppressed MVCT volumes. 

Extensive experiments demonstrate that H3D-MarNet achieves 28.14 dB PSNR and 0.717 SSIM on artifact-affected slices from full dataset, indicating effective metal artifact suppression and anatomical preservation, highlighting its potential for reliable CT modality transformation in clinical radiotherapy workflows.




\end{abstract}

\textit{Accepted for publication at the International Conference on Pattern Recognition (ICPR), 2026.}


\section{Introduction}
\label{sec:intro}

Computed tomography (CT) plays an important role in clinical diagnostics and radiation therapy (RT) planning due to its ability to visualize anatomical structures in high detail. However, metallic implants such as dental fillings or surgical clips introduce severe artifacts caused by beam hardening, photon starvation, and scattering~\cite{Barrett2004}. These distortions appear as streak-like patterns that obscure critical anatomical information~\cite{Park2016,Gjesteby2016}, reducing diagnostic reliability and compromising the accuracy of RT dose calculations~\cite{Giantsoudi2017}.





\begin{figure*}[hbt!]
  \centering

  \begin{minipage}[t]{0.43\textwidth}
    \includegraphics[width=\textwidth]{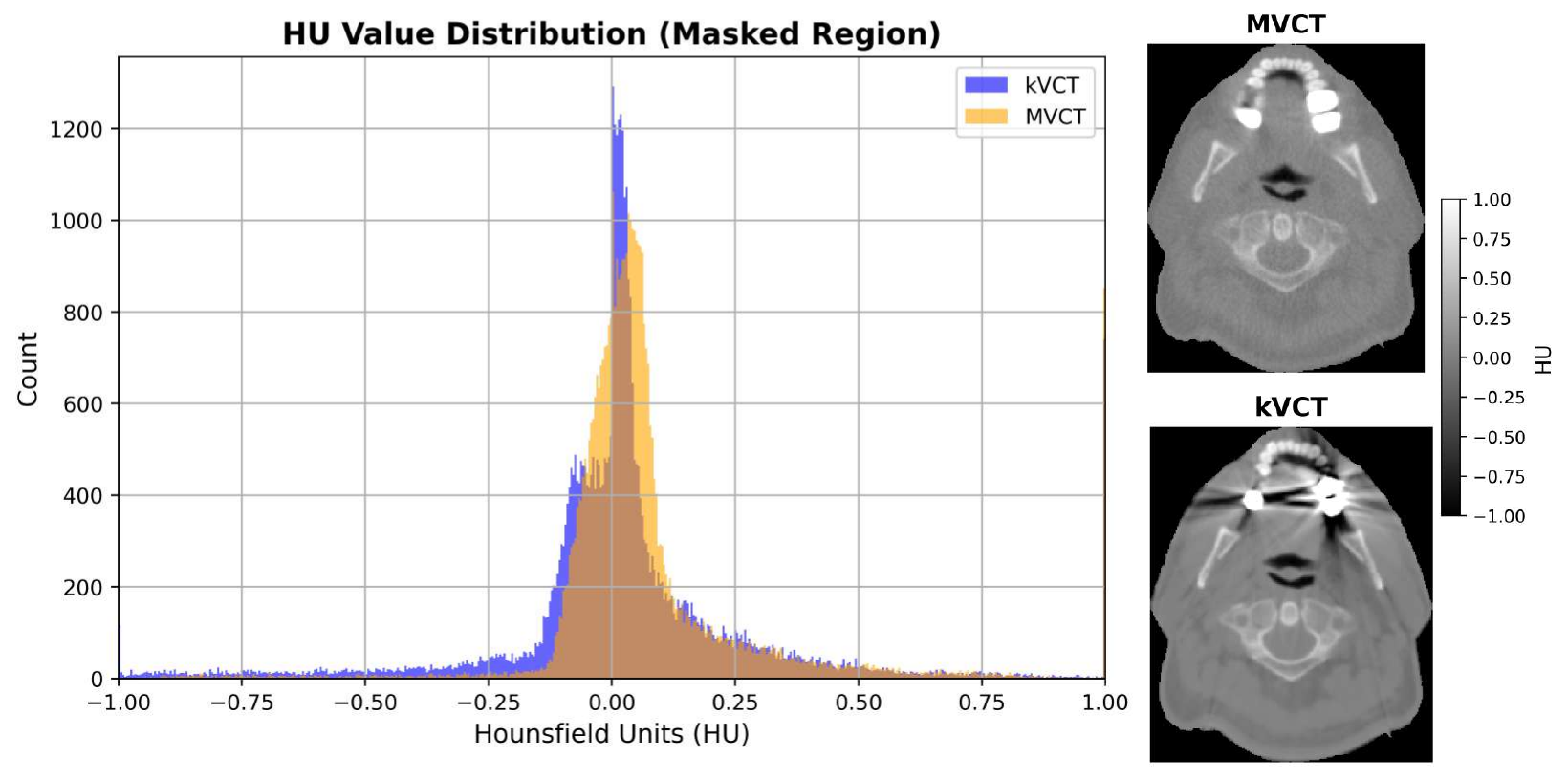}
    \caption{Comparison of HU distributions in normalized kVCT and MVCT slices. 
    kVCT exhibits skewness due to metal artifacts, while MVCT maintains a smoother profile.}
    \label{fig:kvct-mvct-hu-statistics}
  \end{minipage}
  \hfill
  \begin{minipage}[t]{0.55\textwidth}
    \includegraphics[width=\textwidth]{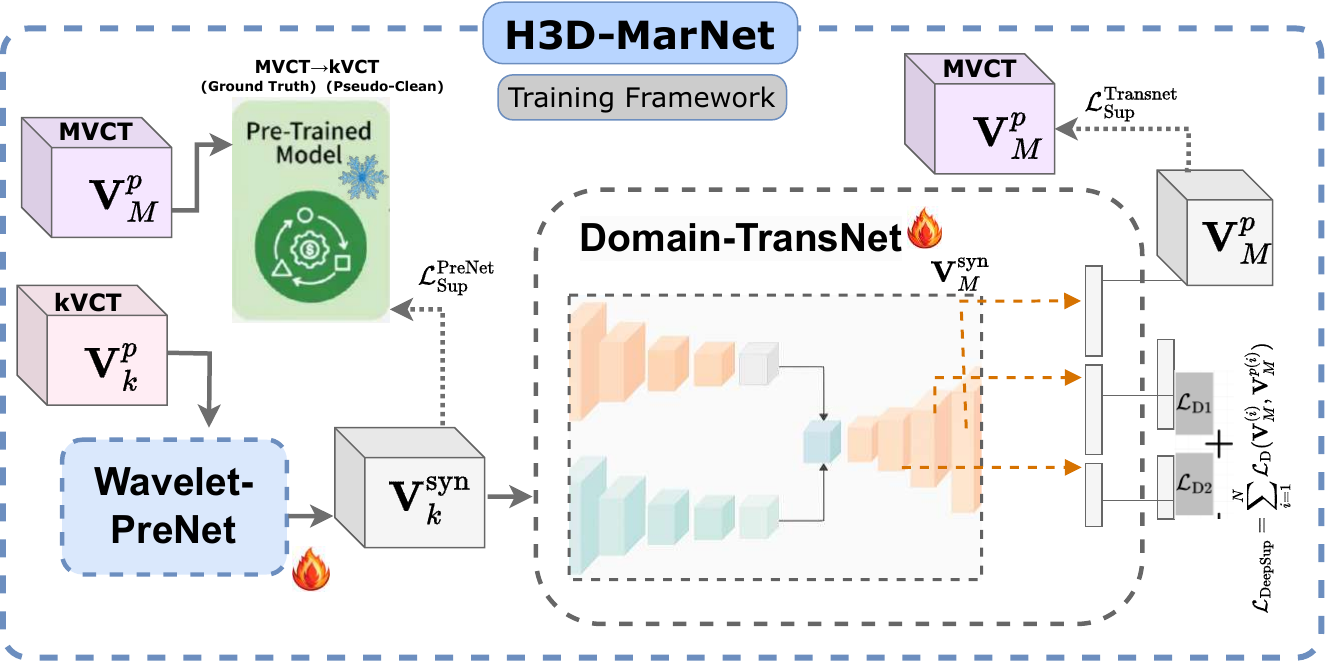}
\caption{Overview of the H3D-MarNet framework. Wavelet-PreNet suppresses artifacts in kVCT inputs, while Domain-TransNet fuses CNN-Transformer features to generate artifact-reduced MVCT outputs.}
    \label{fig:h3d-MarNet-training}
  \end{minipage}

\end{figure*}


To mitigate these effects, metal artifact reduction (MAR) techniques have been widely studied. Traditional image-domain methods have evolved significantly with the adoption of deep learning~\cite{zhang2018residual,Ledig2017,Zhang2018,Ronneberger2015}, which enables data-driven suppression of artifact patterns. Residual CNNs~\cite{Gjesteby2018}, dictionary-based representations, and prior-encoded networks such as DICDNet~\cite{DICDNeT_Wang2022} and ACDNet~\cite{Wang2022Adaptive} have shown promising results. U-Net-based models~\cite{Park2018,rehman2025remards,mardtn-icpr2024} leverage encoder-decoder structures to retain spatial details while reducing streaks. Wang et al.~\cite{PMID:30693351} applied the pix2pix model~\cite{8100115} to reduce metal artifact in the CT image domain, though they may introduce instability or unrealistic textures~\cite{Xu2022}. More recently, Transformer architectures have demonstrated success in visual tasks by capturing long-range dependencies. Vision Transformers (ViT)~\cite{Dosovitskiy2020} and Swin Transformers~\cite{Liu2021} offer global contextual modeling and have been integrated with convolutional backbones to form hybrid networks~\cite{Guo2022}. Such designs combine local feature learning with global attention, which is highly desirable in CT restoration tasks involving both fine textures and widespread artifacts.

Beyond artifact suppression, domain transformation from kilo-Voltage CT (kVCT) to Mega-Voltage CT (MVCT) is particularly relevant for RT workflows. While kVCT offers superior soft-tissue contrast, it is highly susceptible to metal-induced artifacts that distort Hounsfield Unit (HU) values. In contrast, MVCT acquired using higher energy photons (e.g., 3.5~MV fan beams) is inherently more robust to metal and exhibits smoother, more stable HU distributions~\cite{maerz2015megavoltage,michielsen2013novel}.



\begin{figure*}[htb]
\centering
\includegraphics[width=0.99\textwidth]{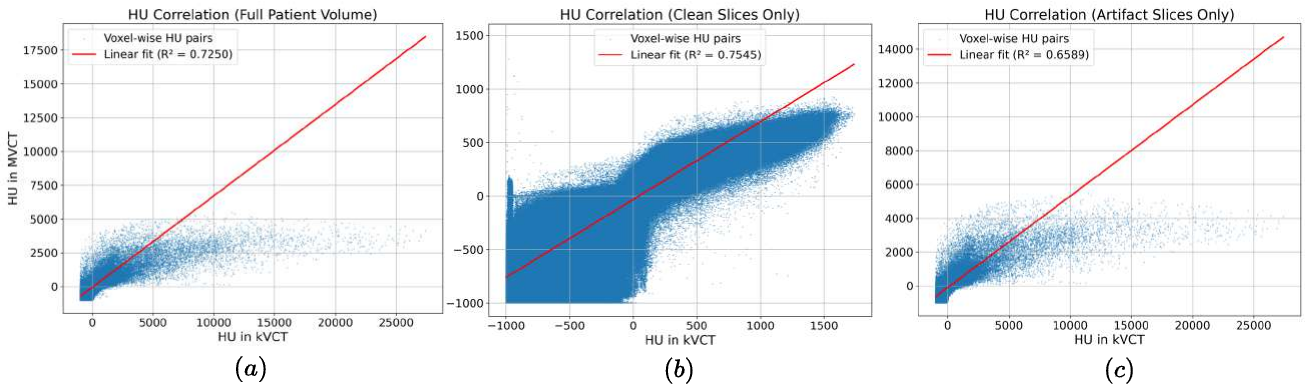}
\caption{Voxel-wise HU correlation between kVCT and MVCT volumes for a representative patient: (a)Entire volume, (b)Clean slices only, and (c)Artifact slices only. The alignment is strongest in clean anatomical regions, validating the feasibility of supervised cross-modal mapping} 
\label{fig:hu_correlation}
\end{figure*}

Learning a supervised kVCT to MVCT mapping enables the synthesis of MVCT-consistent, artifact-resilient images directly from diagnostic kVCT, eliminating the need for additional imaging while preserving HU characteristics critical for RT planning. To support this rationale, Fig.~\ref{fig:kvct-mvct-hu-statistics} shows that kVCT HU histograms exhibit long-tailed skewness due to metal streaking, whereas MVCT distributions remain comparatively symmetric. Voxel-wise HU correlation analysis (Fig.~\ref{fig:hu_correlation}) further reveals strong cross-modal alignment in metal-free slices (\(R^2 = 0.7545\)) and across the full volume (\(R^2 = 0.7250\)), with reduced correlation in artifact-affected regions (\(R^2 = 0.6589\)). These observations justify MVCT as a robust supervisory target for artifact-aware domain transformation.

The MVCT used in our study is acquired using a helical TomoTherapy system equipped with a 3.5 MV fan beam and a portal detector. Due to its higher energy, MVCT is less prone to beam hardening and metal-induced artifacts. Our goal is to synthesize artifact-reduced synthetic MVCT from kVCT inputs, thereby reducing redundant scans and radiation exposure. Clinical evaluation confirmed that synthetic MVCT preserves HU fidelity and dosimetric accuracy across organs-at-risk (OARs) and targets, supporting its suitability for adaptive RT in head and neck cancer.




However, learning such a kVCT to MVCT mapping is challenging due to the simultaneous presence of severe metal artifacts and substantial cross-modal appearance differences. Unified one-stage networks~\cite{rehman2025remards,mardtn-icpr2024} must implicitly solve two tightly coupled but fundamentally different tasks: (i) suppressing high-frequency metal-induced streaks in kVCT, and (ii) reconstructing anatomically consistent MVCT volumes with modality-specific contrast. This task entanglement often limits complete artifact removal and may introduce structural distortions near metal regions. To address this, we adopt an explicit two-stage formulation in H3D-MarNet as illustrated in \figurename~\ref{fig:h3d-MarNet-training}. A wavelet-driven artifact suppression stage first produces an artifact-reduced and structurally coherent kVCT representation, which is then processed by a dedicated cross-modal reconstruction module for MVCT synthesis. This principled task decoupling reduces cross-objective interference, stabilizes optimization, and leads to improved volumetric consistency.


\vspace{0.5em}
\noindent The key contributions of this work are:
\begin{itemize}
    \item A wavelet-based preprocessing module that reduces metal-induced high-frequency artifacts via multi-level decomposition, and attention-guided denoising.
    
    
    


    \item A unified 3D reconstruction architecture that integrates a dual-encoder design (MultiRes CNN and Transformer-based volumetric encoding), a cross-attention fusion module with spatial-channel recalibration, and a progressive decoder with hybrid upsampling and recalibrated skip connections, enabling coherent multi-scale feature refinement and MVCT reconstruction with high anatomical fidelity.
    
    \item Extensive experiments showing state-of-the-art performance in both artifact reduction and CT domain transformation, with improved PSNR, SSIM, and visual quality.
\end{itemize}

\noindent The proposed approach facilitates artifact-aware domain translation from kVCT to MVCT, offering improved image quality for RT planning while reducing additional radiation exposure from repeat scans.


\section{Methodology}
\label{sec:method}

\subsection{Formulation and Objective}

\begin{figure*}[t]
\centering
\includegraphics[width=0.99\textwidth]{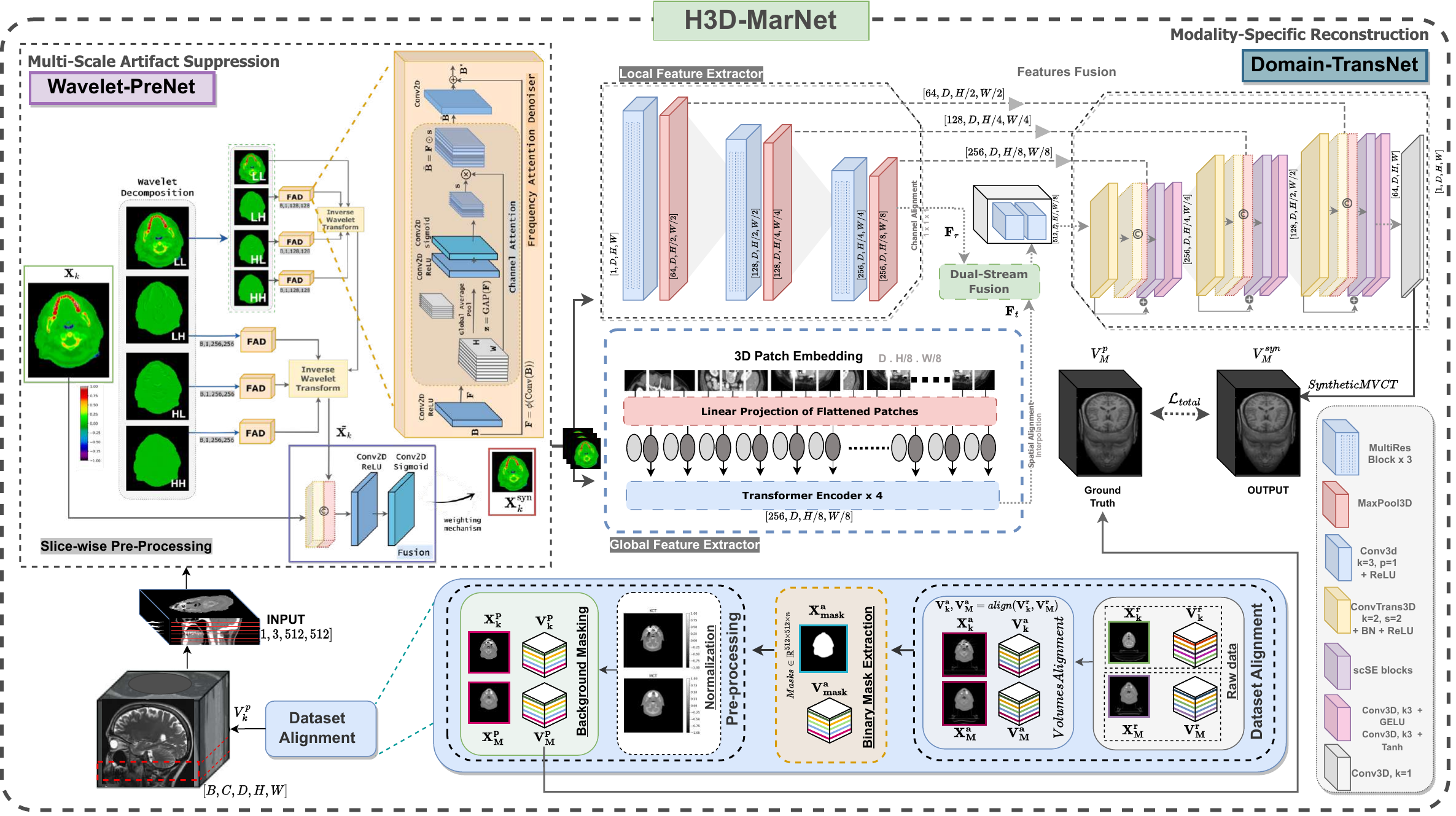}
\caption{Overview of the proposed framework H3D-MarNet, for CT domain transformation with metal artifact reduction.} 
\label{fig:overalarchitecture}
\end{figure*}

Let \( \mathbf{V}_k^p \in \mathbb{R}^{D \times H \times W} \) represent a metal-affected input kVCT volume, where \( D \), \( H \), and \( W \) denote depth, height, and width, respectively. Our goal is to generate an artifact-suppressed MVCT-equivalent volume \( \mathbf{V}_M^{\text{syn}} \in \mathbb{R}^{D \times H \times W} \), preserving anatomical fidelity for clinical use in radiotherapy planning. As illustrated in \figurename\ref {fig:overalarchitecture}, The H3D-MarNet pipeline is modeled as a two-stage transformation:

\begin{equation}
\mathbf{V}_M^{\text{syn}} = f^{\text{TransNet}}_\phi(f^{\text{PreNet}}_\theta(\mathbf{V}_k^p)),
\label{eq:prob-formulation}
\end{equation}

where \( f^{\text{PreNet}}_\theta \) is the wavelet-based preprocessing module (Wavelet-PreNet) that performs early artifact suppression, and \( f^{\text{TransNet}}_\phi \) is the domain transformation network (Domain-TransNet) composed of a dual-path encoder-decoder. The intermediate output from PreNet is denoted by \( \mathbf{V}_k^{\text{syn}} = f^{\text{PreNet}}_\theta(\mathbf{V}_k^p) \), and the final MVCT prediction is \( \mathbf{V}_M^{\text{syn}} = f^{\text{TransNet}}_\phi(\mathbf{V}_k^{\text{syn}}) \), supervised against the true MVCT target \( \mathbf{V}_M^p \).

To facilitate training without requiring paired clean kVCT ground truth, we leverage a pretrained teacher network to generate pseudo-clean kVCT volumes \( \mathbf{V}_k^{\text{clean}} \) from cleaned aligned MVCT-kVCT pairs. This enables the first stage to be trained, while the second stage is directly deep supervised using the processed true MVCT volume \( \mathbf{V}_M^p \). This dual-stage supervision strategy ensures consistent anatomical alignment and domain adaptation.


\subsection{Proposed Framework}

The proposed H3D-MarNet is a two-stage architecture designed to progressively refine metal-affected kVCT volumes. It begins with a frequency-aware preprocessing module for early suppression of high-frequency artifact components, followed by a hybrid dual-encoder 3D reconstruction network for structure-aware refinement and domain transformation. A schematic overview is shown in Fig.\ref {fig:overalarchitecture}.

\subsubsection{Multi-Scale Artifact Suppression via Wavelet-PreNet:}

To suppress high-frequency metal artifacts prior to 3D processing, we introduce module \textbf{Wavelet-PreNet($\mathcal{M}_{\text{Pre}}$)}, a lightweight frequency-aware preprocessing module applied slice-wise to the kVCT volume. Each axial slice $\mathbf{X}_{k} \in \mathbb{R}^{H\times W}$ is first decomposed using a fixed Haar transform $\mathcal{W}(\mathbf{X}_{k})$ into sub-bands $\mathcal{B}^1 = \{\mathbf{B}^1_{\mathrm{LL}}, \mathbf{B}^1_{\mathrm{LH}}, \mathbf{B}^1_{\mathrm{HL}}, \mathbf{B}^1_{\mathrm{HH}}\}$, where the high-frequency components predominantly encode streak artifacts.
To further separate low-frequency anatomical content from residual artifact leakage, a second-level Haar decomposition $\mathcal{W}(\mathbf{B}^1_{\mathrm{LL}})$ is applied, yielding $\mathcal{B}^2 = \{\mathbf{B}^2_{\mathrm{LL}},\allowbreak
\mathbf{B}^2_{\mathrm{LH}},\allowbreak
\mathbf{B}^2_{\mathrm{HL}},\allowbreak
\mathbf{B}^2_{\mathrm{HH}}\}$. All subsequent denoising and reconstruction steps follow the same formulation for both levels.


Each sub-band $\mathbf{B} \in \mathcal{B}^1 \cup \mathcal{B}^2$ is refined using a Frequency-Aware Denoiser (FAD) inspired by squeeze-and-excitation~\cite{hu2018senet} attention. The denoiser extracts localized features via $\mathbf{F}=\mathrm{ReLU}(\mathrm{Conv}_{3\times3}(\mathbf{B}))$, followed by global average pooling to obtain channel-wise descriptors $\mathbf{z}=\mathrm{GAP}(\mathbf{F})$. The excitation step computes a two-layer gating function $\mathbf{s}=\sigma(\mathbf{W}_2\,\delta(\mathbf{W}_1\mathbf{z}))$, where $\delta$ and $\sigma$ denote ReLU and sigmoid activations, and the resulting weights recalibrate the band as $\tilde{\mathbf{B}}=\mathbf{F}\odot\mathbf{s}$. A residual refinement $\mathbf{B}^\ast=\mathbf{B}+\mathrm{Conv}(\tilde{\mathbf{B}})$ yields the denoised sub-band. Unlike traditional SE blocks operating in the spatial domain, our formulation enhances frequency-localized features, allowing for fine-grained denoising in a multiscale context.


The cleaned components are recomposed via inverse Haar reconstruction to produce a wavelet-refined slice
$\tilde{\mathbf{X}}_{k} = \mathcal{W}^{-1}(\mathbf{B}^{\ast}_{\mathrm{LL}}, \mathbf{B}^{\ast}_{\mathrm{LH}}, \mathbf{B}^{\ast}_{\mathrm{HL}}, \mathbf{B}^{\ast}_{\mathrm{HH}})$.
To estimate a spatial mixing map $\mathbf{w}$, the raw slice $\mathbf{X}_{k}$ and its wavelet-refined counterpart $\tilde{\mathbf{X}}_{k}$ are first concatenated channel-wise and passed through a lightweight fusion module consisting of a Conv–ReLU followed by a Conv–Sigmoid layer. The final artifact-suppressed slice is then obtained via spatially adaptive weighted fusion as; $\mathbf{X}_k^{\mathrm{syn}}=\mathbf{w}\odot\mathbf{X}_{k}+(1-\mathbf{w})\odot \tilde{\mathbf{X}_{k}}$.
These artifact-suppressed slice are stacked to reconstruct the artifact-suppressed volume \( \mathbf{V}_k^{\text{syn}} \in \mathbb{R}^{1 \times D \times H \times W} \). This volume serves as input to the subsequent 3D domain transformation stage.


\subsubsection{Modality-Specific Reconstruction via Domain-TransNet:}

The artifact-suppressed kVCT volume \( \mathbf{V}_k^{\text{syn}} \in \mathbb{R}^{D \times H \times W} \) is processed by a dual-path 3D encoder. The convolutional branch \( \mathcal{M}_{\text{CNN}} \) follows a 3D U-Net-style design and employs stacked residual (MultiRes) blocks to extract hierarchical spatial features, yielding
\( F_r = \mathcal{M}_{\text{CNN}}(\mathbf{V}_k^{\text{syn}}) \in \mathbb{R}^{256 \times D \times H/8 \times W/8} \).
Each MultiRes block applies a residual mapping
\( \mathbf{h}_{l+1} = \mathbf{h}_l + \mathcal{F}_{\text{MR}}(\mathbf{h}_l) \),
where \( \mathcal{F}_{\text{MR}}(\cdot) \) consists of stacked \(3\times3\times3\) convolutional layers with nonlinear activation. Three such residual blocks are applied per encoder level before spatial downsampling. Feature channels increase from 64 to 128 and then to 256, while spatial resolution is reduced by a factor of 8.

In parallel, the transformer-based branch \( \mathcal{M}_{\text{Trans}} \) models long-range volumetric dependencies via patch-wise token representations. The volume is partitioned into \( N \) non-overlapping 3D patches of size \( P \times P \times P \), each flattened into a token \( x_i^p \in \mathbb{R}^{P^3 C} \). Tokens are embedded using a learnable projection \( E \in \mathbb{R}^{(P^3 C)\times d} \) and positional encoding \( E_{\text{pos}} \in \mathbb{R}^{N\times d} \), forming
\( Z_0 = [x_1^p E;\dots;x_N^p E] + E_{\text{pos}} \).
A 4-layer Transformer encoder processes the sequence, which is reshaped to obtain
\( F_t \in \mathbb{R}^{256 \times D \times H/8 \times W/8} \).

The features \( F_r \) and \( F_t \) are fused using a dual-attention module \( \mathcal{M}_{\text{Fuse}} \) as
\( F_f = \phi\!\big( \alpha_s(\psi(F_r \oplus F_t)) \odot \alpha_c(\psi(F_r \oplus F_t)) \big) \),
where \( \psi(\cdot) \) is a \(1\times1\times1\) convolution for feature alignment, \( \alpha_s(\cdot) \) and \( \alpha_c(\cdot) \) denote spatial and channel attention, and \( \phi(\cdot) \) is a bottleneck composed of stacked \(3\times3\times3\) convolutions. The fused representation \( F_f \) is forwarded to the decoder.


The decoder \( \mathcal{M}_{\text{Dec}} \) reconstructs the MVCT-like volume \( \mathbf{V}_M^{\text{syn}} \) through three progressive upsampling stages with feature recalibration. Each stage applies a transposed convolution followed by two \(3\times3\times3\) convolutional refinements with GELU activation, while encoder features are incorporated via selective skip recalibration using attention gating.

To enable deep supervision, the feature maps after each upsampling stage \( \{\mathbf{F}_1, \mathbf{F}_2, \mathbf{F}_3\} \) are independently mapped to MVCT predictions using identical prediction heads, defined as \( \mathbf{V}_M^{(i)} = \mathrm{Conv}_{1\times1\times1}(\mathrm{Tanh}(\mathrm{Conv}_{1\times3\times3}^{\text{GELU}}(\mathbf{F}_i))) \), \( i \in \{1,2,3\} \). Here, \( \mathbf{V}_M^{(1)} \) and \( \mathbf{V}_M^{(2)} \) serve as auxiliary outputs for deep supervision, while \( \mathbf{V}_M^{(3)} \) corresponds to the final full-resolution MVCT prediction \( \mathbf{V}_M^{\text{syn}} \).


\subsubsection*{Loss Functions}

Our proposed framework employs a multi-stage supervision strategy to guide learning in both stages. In artifact suppression stage, the artifact-reduced output \( \mathbf{V}_k^{\text{syn}} = f^{\text{PreNet}}_\theta(\mathbf{V}_k^p) \) is supervised using a pseudo-clean kVCT volume \( \mathbf{V}_k^{\text{clean}} \), generated by a teacher model trained on clean kVCT-MVCT pairs. In the second stage, the synthesized MVCT prediction
\( \mathbf{V}_M^{\text{syn}} = \allowbreak f^{\text{TransNet}}_\phi(\mathbf{V}_k^{\text{syn}}) \)
is supervised using the real MVCT target \( \mathbf{V}_M^p \).
To enhance convergence and gradient flow, auxiliary supervision is imposed on intermediate decoder outputs \( \mathbf{V}_M^{(i)} \) at multiple scales.

The supervision loss for each stage is formulated as a weighted combination of spatially-aware L1 loss (\( \mathcal{L}_1^w \)), SSIM loss (\( \mathcal{L}_{\text{SSIM}} \)), and perceptual loss (\( \mathcal{L}_{\text{Percep}} \)). The combined supervision loss is given by:
\begin{equation}
\mathcal{L}_{\text{Sup}} = \lambda_1 \mathcal{L}_1^w + \lambda_s \mathcal{L}_{\text{SSIM}} + \lambda_p \mathcal{L}_{\text{Percep}},
\end{equation}
where \( \lambda_1 = 1.0 \), \( \lambda_s = 0.5 \), and \( \lambda_p = 0.5 \) are empirically chosen weights. Specifically, \( \mathcal{L}_1^w \) penalizes pixel-wise spatial errors with metal-aware weighting, \( \mathcal{L}_{\text{SSIM}} \) promotes local structural preservation, and \( \mathcal{L}_{\text{Percep}} \) enforces perceptual fidelity with features from a pretrained VGG-16 network.

To enforce multi-scale consistency, deep supervision is applied to each intermediate decoder prediction:
\begin{equation}
\mathcal{L}_{\text{DeepSup}} = \sum_{i=1}^{N} \mathcal{L}_{\text{D}}(\mathbf{V}_M^{(i)}, \mathbf{V}_M^{p(i)}),
\label{eq:deep-supervision}
\end{equation}
where \( \mathbf{V}_M^{p(i)} \) denotes the ground truth MVCT downsampled to the resolution of the \( i \)-th decoder output, and \( \mathcal{L}_{\text{D}} \) is the L1 loss at that scale.

The final objective aggregates supervision across both stages along with deep supervision:
\begin{equation}
\mathcal{L}_{\text{total}} =
\lambda_a \cdot \mathcal{L}_{\text{Sup}}^{\text{PreNet}} +
\lambda_b \cdot \mathcal{L}_{\text{Sup}}^{\text{TransNet}} +
\lambda_c \cdot \mathcal{L}_{\text{DeepSup}},
\label{eq:total-loss}
\end{equation}

where \( \lambda_a = 1.0 \), \( \lambda_b = 1.0 \), and \( \lambda_c = 0.5 \) are empirically selected to balance artifact reduction, domain transformation, and auxiliary deep supervision, respectively.


This composite loss drives the model to learn an end-to-end mapping from the input volume $\mathbf{V}_k^p$ to the synthesized MVCT volume $\mathbf{V}_M^{\text{syn}}$, enforcing spatial, structural, and perceptual alignment with the ground-truth anatomy.


\section{Experiments and Results}

\begin{figure}[t]
  \centering
  \includegraphics[width=0.4\linewidth]{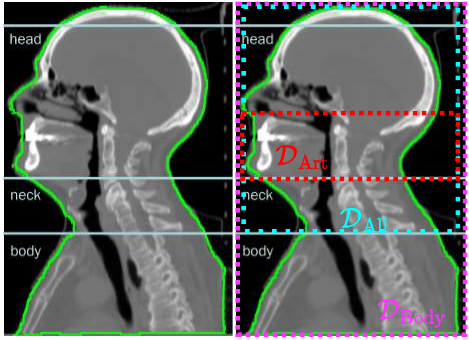}
  \caption{(Left) Sagittal CT slice with anatomical regions (head, neck, body) highlighted. (Right) Dataset distribution across full CT slices (\( \mathcal{D}_\text{All} \)) and artifact-contaminated slices (\( \mathcal{D}_\text{Art} \)).}
  \label{fig:regionsAndBodySegmenation}
\end{figure}

\subsection{Datasets Acquisition and Evaluation Metric}

Unaligned kVCT and MVCT datasets were obtained from \textit{IRCCS}\footnote{Centro di Riferimento Oncologico di Aviano IRCCS, Via F. Gallini 2, Aviano (PN), 33081, Italy.}, comprising 52 patients who underwent IMRT for oropharyngeal or nasopharyngeal cancer. kVCT scans were acquired using a Toshiba Aquilion LB scanner (120 kVp, 2-5 mm slice thickness, 1.07-1.17 mm pixel size), while MVCT images were obtained via helical tomotherapy (Hi-Art II system) using a 3.5 MV imaging beam with a slice thickness of 2-5 mm and a pixel size of 0.75 mm. Due to differences in spatial resolution and acquisition geometry, the kVCT and MVCT volumes were aligned prior to further processing using our in-house multi-modal alignment framework, ensuring consistent anatomical correspondence across modalities. Each volume was manually categorized into three regions (see Fig.~\ref{fig:regionsAndBodySegmenation}(left)): head, neck, and body. 
The study focuses on artifact removal in the dental region, excluding body slices. For quantitative evaluation, PSNR/SSIM are computed within the region of interest (ROI) using segmentation masks. Since MVCT is not entirely artifact-free, qualitative analysis and clinician evaluations supplement the assessment. Artifacts are defined as intensity values exceeding 2000 HU (kVCT) and 1000 HU (MVCT) based on prior studies \cite{kim2022metal,liugang2016metal}. We construct two datasets (see Fig.~\ref{fig:regionsAndBodySegmenation}(right)): $\mathcal{D}_\text{All}$, containing all slices up to the neck (artifact and non-artifact), and $\mathcal{D}_\text{Art}$, comprising only artifact-contaminated slices, which constitute 14.78\% of the dataset, furthermore, both datasets are patient-wise split into 70\% training and 30\% testing.

\subsection{Implementation Details}

The models were trained on an Intel Xeon Server with 188GB of RAM and an Nvidia A100 GPU, with a batch size of $4$, for $200$ epochs, using early stopping with a patience of 15 epochs. The depth for 3D input is $3 \times 512 \times 512$, and produces a output of the same dimensions. Model is optimized using AdamW with the learning rate initialized at $0.0001$ and halved every 20 epochs. Weight decay is set to $5e^{-4}$. To increase the training set, data augmentation was applied using Albumentations \cite{Buslaev_2020}. Data augmentation included horizontal flip with a probability of 0.5, and shift, scale, and rotate transformations with a probability of 0.8 (\textit{shift\_limit}$=0.0625$, \textit{scale\_limit}$=0.1$, \textit{rotate\_limit}$=5$).



\begin{figure*}[htb]
\centering
\includegraphics[width=\textwidth]{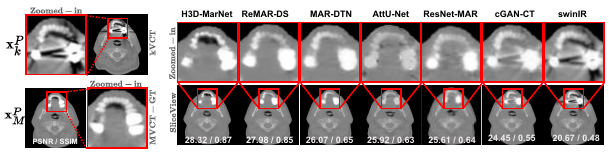}
\caption{Illustration of the reconstruction of MVCT, using various models. All networks were trained on the artifact dataset, highlighting their performance in artifact reduction and image reconstruction. Quantitative metrics for each reconstruction are provided, including PSNR and SSIM.} 
\label{fig:comparison_model_slice_grid}
\end{figure*}


\begin{figure*}[h!]
\centering
\includegraphics[width=1.0\textwidth]{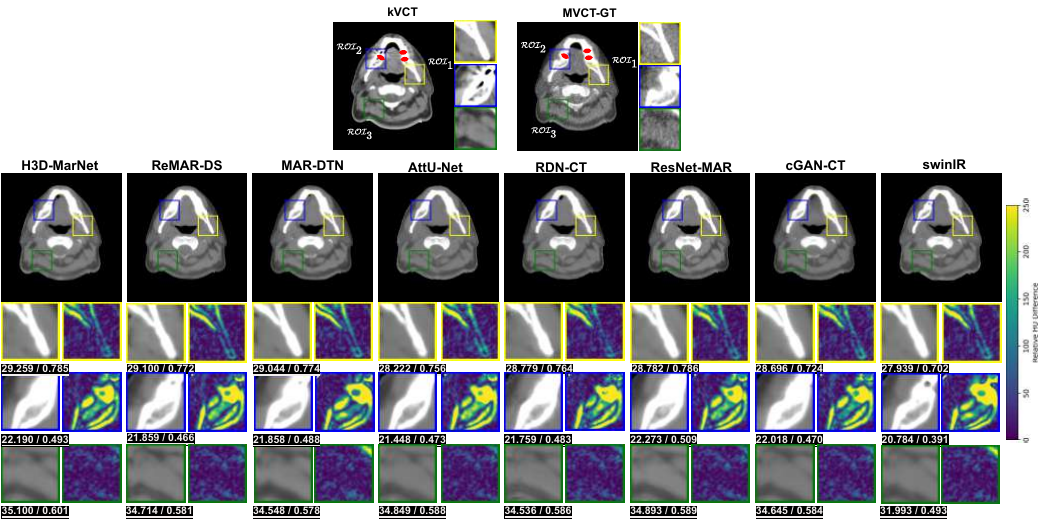}
\caption{ROI-wise quantitative and visual evaluation of MVCT reconstructions across all models. Top: reconstructed slices with ROI annotations. Middle: ROI patches and corresponding relative HU-difference maps (0–250 HU). Bottom: PSNR and SSIM metrics computed against the ground-truth MVCT. The selected ROIs emphasize metal-adjacent bone (\textcolor{yellow!80!black}{yellow}), streak-affected soft tissue (\textcolor{blue!80!black}{blue}), and homogeneous tissue (\textcolor{green!60!black}{green}).} 
\label{fig:HU-ROI-Grid}
\end{figure*}



\begin{table*}[t]

    \caption{Quantitative performance comparison on $\mathcal{D}_{\text{Art}}$ (artifact slices) and $\mathcal{D}_{\text{All}}$ (entire test set). Metrics include PSNR and SSIM. Training time is computed for the $\mathcal{D}_{\text{All}}$ per epoch; reconstruction time refers to a 170-slice patient volume. \textbf{Bold} indicates the best result per column. NMAR and FSMAR are non-learned baselines.}
    \centering
    \resizebox{12cm}{!}{
    \begin{tabular}{lcccccccc}
        \Xhline{1.5pt}
        \multirow{2}{*}{Model} 
        & PSNR ↑ & SSIM ↑ & PSNR ↑ & SSIM ↑ 
        & Params ↓ & MACs ↓ & Train Time & Recon Time \\
        & $\mathcal{D}_{\text{Art}}$ & $\mathcal{D}_{\text{Art}}$ 
        & $\mathcal{D}_{\text{All}}$ & $\mathcal{D}_{\text{All}}$ 
        & (M) & (G) & (s/epoch) & (s) \\
        \Xhline{1.5pt}

        \rowcolor{lightergray} 
        H3D-MarNet (ours) & \textbf{28.02} & \textbf{0.71} & \textbf{28.14 (30.26)} & \textbf{0.717 (0.762)} & 22.6 & 184.12 & 684.1 & 6.23 \\

        ReMAR-DS~\cite{rehman2025remards} & 27.67 & 0.71 & 27.85 (30.69) & 0.72 (0.76) & 11.55 & 650.0 & 256.55 & 3.56 \\
        MAR-DTN~\cite{mardtn-icpr2024} & 27.46 & 0.69 & 27.05 (30.02) & 0.69 (0.73) & 1.88 & 116.86 & 65.32 & 3.05 \\
        AttU-Net~\cite{wang2022attu-net} & 27.09 & 0.69 & 26.69 (30.12) & 0.68 (0.73) & 34.83 & 1922.88 & 735.0 & 5.92 \\
        ResNet-MAR~\cite{ledig2017photo} & 26.85 & 0.67 & 27.78 (30.45) & 0.72 (0.75) & 640.39 & 168.18 & 256.55 & 7.22 \\
        RDN-CT~\cite{zhang2018residual} & 26.71 & 0.67 & 25.61 (28.36) & 0.59 (0.65) & 13.76 & 3605.93 & 2,647 & 52.3 \\
        cGAN-CT~\cite{PMID:30693351} & 26.31 & 0.64 & 26.36 (28.70) & 0.63 (0.69) & 64.66 & 208.31 & 80.02 & 3.75 \\
        SwinIR~\cite{liang2021swinir} & 25.18 & 0.63 & 25.24 (29.00) & 0.60 (0.66) & 11.8 & 325.03 & 2774.76 & 47.27 \\
        NMAR~\cite{meyer2010normalized} & 23.16 & 0.49 & 23.16 (27.35) & 0.49 (0.60) & -- & -- & -- & 1332.5 \\
        FSMAR~\cite{meyer2012fsmar} & 22.95 & 0.51 & 22.95 (27.07) & 0.51 (0.63) & -- & -- & -- & 1337.2 \\

        \Xhline{1.5pt}
    \end{tabular}
    }

    \label{tab:sota-metrics-comparison}
    \vspace{-0.3cm}
\end{table*}


\begin{table}[t]

\caption{Ablation of individual modules in H3D-MarNet. 
    $\mathcal{M}_{\text{Pre}}$, $\mathcal{M}_{\text{CNN}}$, 
    $\mathcal{M}_{\text{Trans}}$, $\mathcal{M}_{\text{Fuse}}$, 
    and $\mathcal{M}_{\text{Dec}}$ denote the Preprocessing module, 
    CNN encoder branch, Transformer encoder branch, Dual-Stream Fusion 
    module, and Enhanced Decoder, respectively. 
    Only v1 includes $\mathcal{M}_{\text{Pre}}$, while v2-v5 are trained 
    without the preprocessing module. 
    PSNR/SSIM are reported on $\mathcal{D}_{\text{Art}}$ and 
    $\mathcal{D}_{\text{All}}$; values in parentheses reflect the full 
    test set. \textbf{Bold} marks best performance.}
    \centering
    \tiny
    \resizebox{11cm}{!}{
    \setlength{\tabcolsep}{3.5pt}
    \renewcommand{\arraystretch}{1.5}
    \resizebox{\linewidth}{!}{
    \begin{tabular}{lccccccc}
        \Xhline{1.0pt}
        \multirow{2}{*}{Ver.} 
        & $\mathcal{M}_{\text{Pre}}$ 
        & $\mathcal{M}_{\text{CNN}}$ 
        & $\mathcal{M}_{\text{Trans}}$ 
        & $\mathcal{M}_{\text{Fuse}}$ 
        & $\mathcal{M}_{\text{Dec}}$ 
        & $\mathcal{D}_\text{Art}$ 
        & $\mathcal{D}_\text{All}$ \\
        & & & & & 
        & PSNR / SSIM ↑ & PSNR / SSIM ↑ \\
        \Xhline{1.0pt}

        \rowcolor{lightergray}
        v1 & \checkmark & \checkmark & \checkmark & \checkmark & \checkmark 
        & \textbf{28.02 / 0.71} 
        & \makecell{\textbf{28.14 / 0.717} \\ (\textbf{30.26 / 0.762})} \\

        v2 & \ding{55} & \checkmark & \checkmark & \ding{55} & \checkmark 
        & 27.61 / 0.693 
        & \makecell{27.84 / 0.705 \\ (29.72 / 0.744)} \\

        v3 & \ding{55} & \checkmark & \checkmark & \checkmark & \checkmark
        & 27.74 / 0.698 
        & \makecell{27.93 / 0.709 \\ (29.88 / 0.748)} \\

        v4 & \ding{55} & \checkmark & \checkmark & \checkmark & \ding{55} 
        & 27.51 / 0.685 
        & \makecell{27.72 / 0.701 \\ (29.41 / 0.739)} \\

        v5 & \ding{55} & \checkmark & \ding{55} & \ding{55} & \checkmark 
        & 27.48 / 0.668 
        & \makecell{27.67 / 0.692 \\ (29.01 / 0.721)} \\
        \Xhline{1.0pt}
    \end{tabular}
    }
}    
    \label{tab:model-ablation-component}
\end{table}


\begin{table}[t]
\caption{Performance ablation of different loss function combinations. PSNR/SSIM are reported on $\mathcal{D}_\text{Art}$ and $\mathcal{D}_\text{All}$. Parentheses indicate full test set evaluation. \textbf{Bold} indicates best results.}
    \centering
    \tiny
         \resizebox{12cm}{!}{
    \setlength{\tabcolsep}{3.5pt}
    \renewcommand{\arraystretch}{1.5}
    \resizebox{8.5cm}{!}{
    \begin{tabular}{lcccccccc}
        \Xhline{1.0pt}
        \multirow{2}{*}{Loss} 
        & $\mathcal{L}_1$ & $\mathcal{L}_{\textit{SSIM}}$ & $\mathcal{L}_{\textit{MS-SSIM}}$ & $\mathcal{L}_{\textit{MSE}}$ 
        & $\mathcal{L}_{\textit{FFL}}$ & $\mathcal{L}_{\textit{Percep}}$ 
        & $\mathcal{D}_\text{Art}$ & $\mathcal{D}_\text{All}$ \\
        Ver. &  & & & & & 
        & PSNR / SSIM ↑ & PSNR / SSIM ↑ \\
        \Xhline{1.0pt}

        L1 & \checkmark &  &  &  &  &  
        & 27.412 / 0.697 
        & \makecell{27.738 / 0.706 \\ (29.985 / 0.742)} \\

        L2 & \checkmark & \checkmark &  &  &  &  
        & 27.703 / 0.702 
        & \makecell{27.968 / 0.712 \\ (30.016 / 0.750)} \\

        L3 & \checkmark &  & \checkmark &  &  &  
        & 27.659 / 0.705 
        & \makecell{28.052 / 0.717 \\ (30.114 / 0.751)} \\

        L4 & \checkmark &  &  & \checkmark &  &  
        & 27.672 / 0.693 
        & \makecell{27.996 / 0.704 \\ (29.961 / 0.746)} \\

        L5 & \checkmark & \checkmark &  &  & \checkmark &  
        & 27.715 / 0.704 
        & \makecell{28.074 / 0.715 \\ (30.101 / 0.754)} \\

        \rowcolor{lightergray} L6 & \checkmark & \checkmark &  &  &  & \checkmark 
        & \textbf{28.019 / 0.710} 
        & \makecell{\textbf{28.139 / 0.717} \\ (\textbf{30.256 / 0.762})} \\
        \Xhline{1.0pt}
    \end{tabular}
    }
    }
    
    \label{table:ablation_lossfunctions}
\end{table}


\subsection{Quantitative and qualitative results}

\subsubsection{Comparison with State-of-the-Art Methods}

We benchmark our proposed H3D-MarNet against a diverse suite of state-of-the-art (SOTA) methods, including traditional, CNN-based, and Transformer-based approaches. Traditional baselines include NMAR~\cite{meyer2010normalized}, which performs sinogram inpainting via normalization, and FSMAR~\cite{meyer2012fsmar}, which utilizes frequency-split iterative reconstruction. Among CNN-based methods, ResNet-MAR is adapted from SRResNet~\cite{ledig2017photo} with eight residual blocks for direct artifact regression. AttU-Net~\cite{wang2022attu-net} incorporates attention gates into the U-Net backbone to improve skip connection modulation.  SwinIR~\cite{liang2021swinir}, a Transformer model with shifted window attention, is repurposed for CT artifact suppression. MAR-DTN~\cite{mardtn-icpr2024} and ReMAR-DS~\cite{rehman2025remards} perform direct domain transformation from kVCT to MVCT using an encoder–decoder architecture trained on paired scans. Followed the implementation of RDN-CT~\cite{zhang2018residual}, and cGAN-CT~\cite{PMID:30693351} from~\cite{CVPRDuDoNet} is also included for benchmarking.


Quantitative results in Table~\ref{tab:sota-metrics-comparison} show that H3D-MarNet consistently outperforms all baselines across both the full dataset (\( \mathcal{D}_\text{All} \)) and the artifact-specific subset (\( \mathcal{D}_\text{Art} \)). It achieves the highest PSNR (30.256 dB) and SSIM (0.762) on \( \mathcal{D}_\text{All} \), and maintains strong performance on \( \mathcal{D}_\text{Art} \) with 28.139 dB PSNR and 0.717 SSIM. While MAR-DTN is the strongest baseline, it falls short in both artifact removal and structural preservation. \figurename~\ref{fig:comparison_model_slice_grid} presents a visual comparison of outputs with corresponding PSNR and SSIM scores, highlighting the superior anatomical detail and artifact suppression of H3D-MarNet, particularly in high-metal regions. These improvements are reinforced by SSIM gains, indicating enhanced structural fidelity, and further validated by ROI-based HU error analysis in \figurename~\ref{fig:HU-ROI-Grid}.


\subsubsection{Ablation Study}

We conduct two ablation studies to evaluate the contributions of (1) loss function combinations and (2) architectural components within H3D-MarNet.

Table~\ref{table:ablation_lossfunctions} presents the impact of six loss configurations on model performance. We assess various combinations of weighted L1 loss (\( \mathcal{L}_1^w \)), SSIM (\( \mathcal{L}_\text{SSIM} \)), MS-SSIM (\( \mathcal{L}_\text{MS-SSIM} \)), MSE (\( \mathcal{L}_\text{MSE} \)), Focal Frequency Loss (\( \mathcal{L}_\text{FFL} \))~\cite{jiang2021focal}, and Perceptual Loss (\( \mathcal{L}_\text{Percep} \))~\cite{johnson2016perceptual}. Among these, the combination \( \mathcal{L}_1^w + \mathcal{L}_\text{SSIM} + \mathcal{L}_\text{Percep} \) (denoted L6) achieves the best results on both \( \mathcal{D}_\text{Art} \) and \( \mathcal{D}_\text{All} \), offering a strong balance between spatial accuracy and perceptual fidelity. Its superior SSIM highlights improved structural preservation in artifact-prone regions.

We systematically ablate the core components of H3D-MarNet to quantify their individual contributions. As shown in Table~\ref{tab:model-ablation-component}, removing the Wavelet-PreNet preprocessing module yields the largest performance drop, confirming its effectiveness in early artifact suppression. Disabling the dual-stream fusion block (v2) also degrades performance, indicating the benefit of our proposed multi-domain feature integration over simple concatenation. Similarly, replacing the enhanced decoder with a plain U-Net decoder (v4) reduces reconstruction fidelity. Finally, eliminating the Transformer branch (v5) results in the weakest performance among the ablated variants, underscoring the importance of global contextual modeling. The full configuration (v1) achieves the best overall results 28.02~dB / 0.71~SSIM on $\mathcal{D}_\text{Art}$ and 30.26~dB / 0.762~SSIM on $\mathcal{D}_\text{All}$, highlighting the complementary contributions of preprocessing, dual-branch encoding, fusion, and decoder-level recalibration.

These findings confirm that both the tailored supervision and architectural design synergistically enhance artifact reduction and anatomical consistency. Visual comparisons in \figurename~\ref{fig:HU-ROI-Grid} reinforce these improvements through reduced HU errors across multiple ROIs.

\begin{figure*}[htb]
\centering
\includegraphics[width=0.75\textwidth]{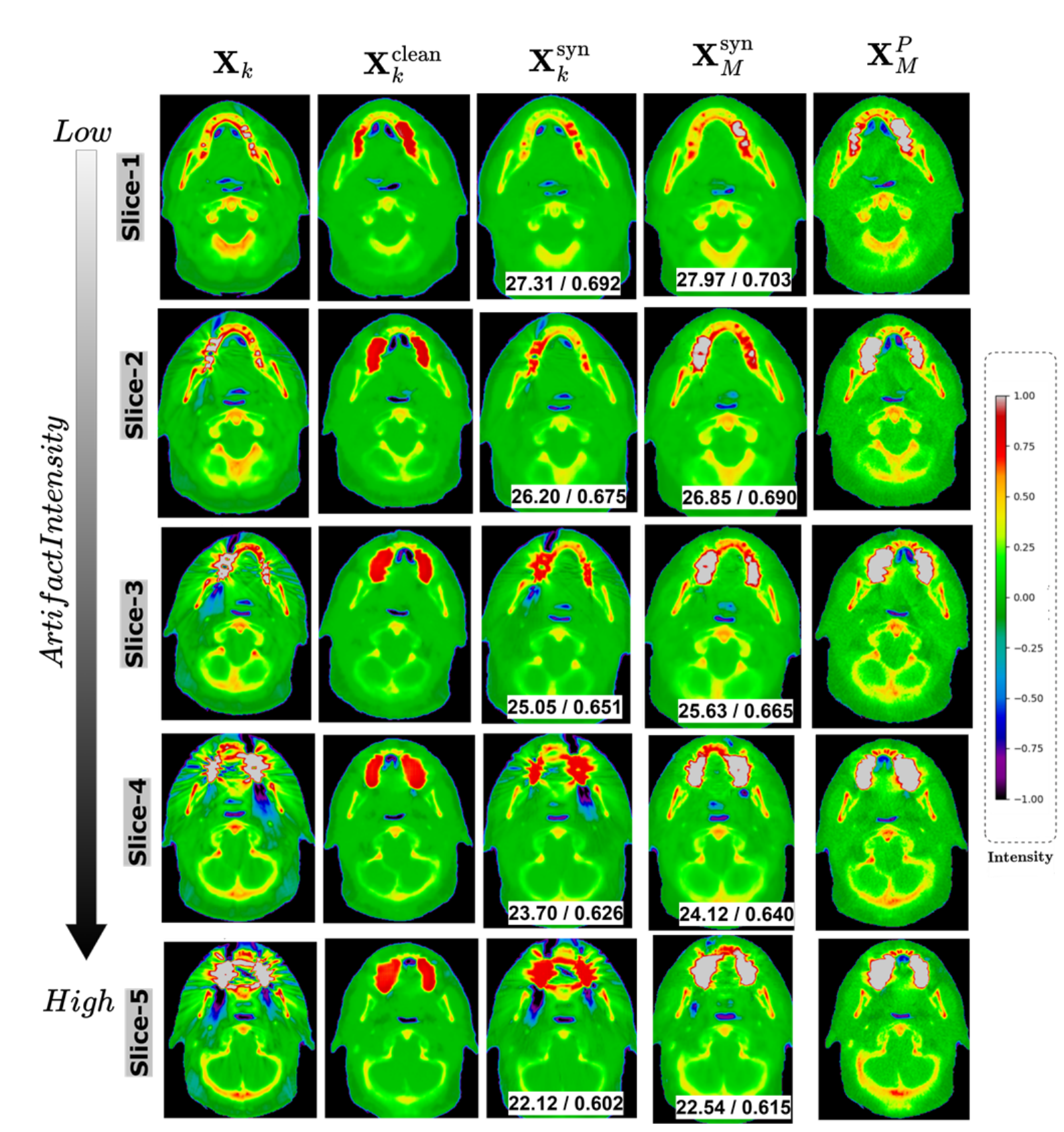}
\caption{Stage-wise reconstruction results from H3D-MarNet across slices with increasing artifact severity. Rows: input kVCT (\( \mathbf{X}_k \)), teacher-predicted kVCT (\( \mathbf{X}_k^{\text{clean}} \)), Stage~1 output (\( \mathbf{X}_k^{\text{syn}} \)), Stage~2 output (\( \mathbf{X}_M^{\text{syn}} \)), and ground-truth MVCT (\( \mathbf{X}_M^p \)). PSNR/SSIM scores indicate artifact suppression and domain alignment quality at each stage.}
\label{fig:grid-artifact-trend}
\end{figure*}

To further evaluate robustness across varying artifact severity, \figurename~\ref{fig:grid-artifact-trend} illustrates a slice-wise analysis. From top to bottom, the input kVCT slices exhibit increasing artifact presence, particularly near high-density implants. The intermediate outputs from the artifact reduction stage show clear suppression of streaks and noise, while the final MVCT predictions maintain anatomical continuity. Notably, even in severe cases (e.g., Slice-5), the model produces reconstructions with stable PSNR and SSIM, reflecting effective generalization and domain alignment.

\subsubsection{Clinical Evaluation}

To assess clinical applicability, radiation oncologists evaluated representative kVCT inputs and synthetic MVCT outputs from both our model and baseline methods. The assessment focused on artifact suppression, soft-tissue visibility, and anatomical fidelity near metal implants and OARs. Experts consistently found that H3D-MarNet preserved structural details with minimal distortion and enhanced tissue contrast, indicating its suitability for radiotherapy planning.

\section{Conclusion}

We proposed H3D-MarNet, a unified framework for joint metal artifact reduction and CT domain transformation from kVCT to MVCT. By integrating wavelet-based preprocessing with a hybrid CNN-Transformer encoder and attention-guided decoding, the proposed method effectively suppresses metal-induced artifacts while preserving fine-grained anatomical structures. Extensive quantitative evaluations demonstrate consistent improvements in PSNR and SSIM on artifact-affected datasets, while qualitative results confirm superior reconstruction fidelity in clinically relevant regions. Although the incorporation of volumetric processing and transformer components increases computational complexity, H3D-MarNet exhibits strong potential for reliable integration into radiotherapy planning workflows. Future work will focus on model compression and improved generalization through self-supervised and cross-domain learning strategies.

\subsubsection*{Acknowledgments}
This work is supported by the Italian Ministry of Health (Ricerca Corrente 2024). We thank the Centro di Riferimento Oncologico di Aviano IRCCS, for providing the datasets and resources used in this study.

\bibliographystyle{splncs04}
\bibliography{bibliography}

\end{document}